\documentclass{article}
\usepackage{ijcai25}

\usepackage{amssymb}
\usepackage{multirow}
\usepackage[table,xcdraw]{xcolor}
\usepackage{enumitem}
\usepackage{caption}
\DeclareCaptionFont{ninept}{\fontsize{9pt}{9pt}\selectfont}
\usepackage[normalem]{ulem}
\usepackage{graphicx}
\usepackage{subcaption} 
\usepackage{amsmath}
\usepackage{amsthm}
\usepackage{times}
\usepackage{soul}
\usepackage{url}
\usepackage[hidelinks]{hyperref}
\usepackage[utf8]{inputenc}
\usepackage[T1]{fontenc}
\usepackage{bm}
\usepackage{savesym}
\usepackage{amsmath}

\usepackage{booktabs}
\usepackage{algorithm}
\usepackage{algorithmic}
\usepackage[switch]{lineno}
\newcommand{\spara}[1]{\smallskip\noindent{\bf #1}}

\title{City-Level Foreign Direct Investment Prediction with Tabular Learning on Judicial Data}

\author{
 Tianxing Wu$^{1,2}$\and
 Lizhe Cao$^1$\and
 Shuang Wang$^{1,2}$\footnote{Corresponding authors.}\and
 Jiming Wang$^3$\and
 Shutong Zhu$^1$\and\\
 Yerong Wu$^1$\And
 Yuqing Feng$^{3*}$
 \\
 \affiliations
 $^1$School of Computer Science and Engineering, Southeast University, Nanjing, China\\
 $^2$Key Laboratory of New Generation Artificial Intelligence Technology and Its Interdisciplinary Applications (Southeast University), Ministry of Education, China\\
 $^3$School of Law, Southeast University, Nanjing, China
 \emails
 \{tianxingwu, caolizhe, shuangwang, jimingwang, shutong\_zhu, yerong.wu, fengyuqing\}@seu.edu.cn
 }

\begin{document}

\maketitle

\begin{abstract}
To advance the United Nations Sustainable Development Goal on promoting sustained, inclusive, and sustainable economic growth, foreign direct investment (FDI) plays a crucial role in catalyzing economic expansion and fostering innovation. Precise city-level FDI prediction is quite important for local government and is commonly studied based on economic data (e.g., GDP). However, such economic data could be prone to manipulation, making predictions less reliable. To address this issue, we try to leverage large-scale judicial data which reflects judicial performance influencing local investment security and returns, for city-level FDI prediction. Based on this, we first build an index system for the evaluation of judicial performance over twelve million publicly available adjudication documents according to which a tabular dataset is reformulated. We then propose a new \textbf{T}abular \textbf{L}earning method on \textbf{J}udicial \textbf{D}ata (\textbf{TLJD}) for city-level FDI prediction. TLJD integrates row data and column data in our built tabular dataset for judicial performance indicator encoding, and utilizes a mixture of experts model to adjust the weights of different indicators considering regional variations. To validate the effectiveness of TLJD, we design cross-city and cross-time tasks for city-level FDI predictions. Extensive experiments on both tasks demonstrate the superiority of TLJD (reach to at least 0.92 \textit{R}\textsuperscript{2}) over the other ten state-of-the-art baselines in different evaluation metrics. 
\end{abstract}

\section{Introduction}
One of the key objectives of the Sustainable Development Goals~\cite{kilanioti2023knowledge} of the United Nations is to ``\textit{promote sustained, inclusive, and sustainable economic growth}''~\cite{UN2015}. However, realizing this objective has become increasingly challenging, especially considering that the past ten years have been marked by a series of global economic recessions, because of heightened geopolitical tensions, regional conflicts, and unforeseen public health crises. According to the estimation of the World Bank, the Ukrainian economy shrinks by 45.1\%, the Russian economy shrinks by 11.2\%, and the economies of emerging markets and developing countries in the Eurasian region contract by 4.1\%~\cite{WorldBank2022}. It is also expected that from 2024 to 2025, the growth rate of nearly 60\% of the world’s economies (accounting for more than 80\% of the global population) will be below the average level of the 2010s~\cite{WorldBank2024}. Amidst the worsening global economic climate, foreign direct investment (FDI), with the positive impact on economic growth, industrial upgrading, and the innovation of cutting-edge technologies, has emerged as a crucial tool to combat local economic recessions. 

As the world’s second largest economy, China has introduced a range of policies and laws to attract FDI in recent years. A significant milestone is the promulgation and enforcement of the Foreign Investment Law of the People’s Republic of China on January 1, 2020, which explicitly states the goal to ``\textit{further expand the opening-up policy, actively promote foreign investment, and protect the legitimate rights and interests of foreign investors}''\footnote{\url{https://www.gov.cn/xinwen/2019-03/20/content\_5375360.htm}}. Additionally, the State Council of the People’s Republic of China issued the Opinions on Further Optimizing the Foreign Investment Environment and Increasing the Attraction of Foreign Investment\footnote{\url{https://www.gov.cn/zhengce/zhengceku/202308/content\_6898049.htm}} on August 13, 2023, further emphasizing the need to create a world-class business environment that is internationalized, law-governed, and market-oriented. 

Despite the increasing significance of FDI in China’s economic strategy, decision-making regarding city-level FDI is fraught with uncertainties. With a lack of capability to predict city-level FDI, it is simply impractical for local governments to make effective and efficient decisions that maximize the benefits and minimize the risks associated with incoming FDI. Economists and computer scientists have conducted extensive studies ~\cite{rw_fdi_ml_financial,rw_fdi_evlution_ml,lassofdi,rw_fdi_ann_qatar,rw_fdi_ann_europe,rw_fdi_ann_Hungarian_counties,rw_fdi_unctad} to improve the prediction accuracy and uncover the determinants of FDI. These studies rely heavily on economic data (e.g., GDP) from official statistics, which could be prone to manipulation by statistical agencies~\cite{ecnomicdata}. This will cause inaccurate city-level FDI prediction, potentially misleading policymakers for local governments. 

To address this issue, we propose to leverage judicial data to predict city-level FDI. Such judicial data are over twelve million publicly available China's adjudication documents, which reflect local judicial performance through large-scale individual cases, offering a more transparent, verifiable, and reliable information source. To systematically evaluate local judicial performance, we build an index system that contains 380 indicators categorized into four types. According to this index system, we first transform adjudication documents into a structured tabular dataset. We then propose a new \textbf{T}abular \textbf{L}earning method on \textbf{J}udicial \textbf{D}ata (\textbf{TLJD}) to predict city-level FDI. In TLJD, transformer layers with arithmetic attention are utilized to encode indicators as embeddings incorporating row features and column features within our built tabular dataset, and a mixture of experts (MoE) model is employed to predict city-level FDI. Considering potential scenarios in the application of city-level FDI prediction, we finally design two evaluation tasks (i.e., estimating missing historical FDI data for specific cities and forecasting future FDI for each city) in the experiments, and the results show the superiority of TLJD compared with the state-of-the-art baselines in different evaluation metrics. 

\spara{Contributions.} The main contributions are summarized as: 
\begin{itemize}[topsep=5pt,leftmargin=1em]
    \item We propose to use judicial data to predict city-level FDI, which is the first work trying to mine adjudication documents for FDI prediction and provides a reliable and practical alternative to traditional economic data based predictions, which may be inaccurate due to data manipulation.  
    \item We build an index system containing 380 indicators of four types for judicial performance evaluation. By calculating the indicator values of each city in a specific year, we build a tabular judicial dataset for city-level FDI prediction. 
    \item We present a new tabular learning method TLJD for city-level FDI prediction. It emphasizes both the feature similarities and sample similarities by fusing the column features and row features in encoding indicators, and fully considers the regional variations on how judicial performance influences FDI by employing an MoE model to dynamically generate weights of indicators.
    \item  We conduct comprehensive experiments on both designed evaluation tasks to validate the effectiveness of our method TLJD. Experimental results not only demonstrate that TLJD outperforms the state-of-the-art baselines in most situations, but also reflect the practicality of using judicial data for city-level FDI prediction.    
\end{itemize}

\section{Related Work}
\subsection{FDI Prediction}
FDI prediction is to predict the unknown FDI value of a region over a given period, and it is a long-standing research topic due to its significance in economic policymaking~\cite{lassofdi}. Previous works apply traditional statistical models~\cite{rw_fdi_arima1,rw_fdi_arima4} to predict FDI, but the accuracy is always unsatisfied since the real data distributions do not satisfy model assumptions. Recently, many works have employed machine learning techniques to develop more effective FDI prediction methods~\cite{rw_fdi_evlution_ml,lassofdi,rw_fdi_lstm,rw_fdi_lstm2,rw_fdi_ann_qatar,rw_fdi_ann_europe,rw_fdi_ann_Hungarian_counties}. For example, an adaptive Lasso grey model~\cite{lassofdi} is proposed to eliminate less important features and alleviate overfitting in FDI prediction. A neural network~\cite{rw_fdi_ann_qatar} is used to model complex relationships among economic features and predict the FDI of Qatar. To capture long-term dependencies in FDI data, a long short-term memory network~\cite{rw_fdi_lstm} is applied in the FDI prediction for Morocco based on various types of data. 

All of the existing studies on FDI prediction rely on economic data, which could be prone to manipulation by statistical agencies~\cite{ecnomicdata}, making the prediction results less reliable. Additionally, the majority of these studies focus on country-level FDI, which cannot effectively help local governments in decision making on economic development. In this paper, we aim to use judicial data extracted from objective large-scale individual cases to predict city-level FDI, providing a reliable and practical technical path of FDI prediction.

\subsection{Tabular Learning}
Tabular learning~\cite{2024review_tabular1} refers to learning on the tabular data organized in a structured table format to solve regression or classification tasks. The judicial data we used in this paper is transformed to a tabular dataset, on which we learn our TLJD for city-level FDI prediction, and TLJD is technically treated as a tabular learning task. This is why tabular learning is relevant to our work. Traditional machine learning methods such as random forest~\cite{2001randomforest} and GBDTs~\cite{2001gbdt} are widely used in tabular learning, as they are quick to train while maintaining competitive performance~\cite{review2}. Recently, deep tabular learning models~\cite{review2} are proposed due to their ability of capturing high-order feature interactions. For example, AutoInt~\cite{song2019autoint} proposes a multi-head self-attentive neural network with residual connections to explicitly model the feature interactions for tabular learning. SAINT~\cite{2021saint} is a specialized tabular learning architecture which projects all categorical and continuous features into a vector space, and applies a hybrid attention mechanism to boost learning performance. FT-Transformer~\cite{2021fttransformer} adopts a stack of transformer layers to feature embedding learning for prediction tasks. AMFormer~\cite{2024amformer} designs a modified transformer architecture enabling arithmetical feature interactions in the tabular learning process. However, such methods cannot be well applied in the scenario of city-level FDI prediction since they do not effectively model the indicators of judicial performance in the tabular dataset, which can be solved by our method TLJD.

\begin{figure}[t]
    \centering
    \includegraphics[width=\columnwidth]{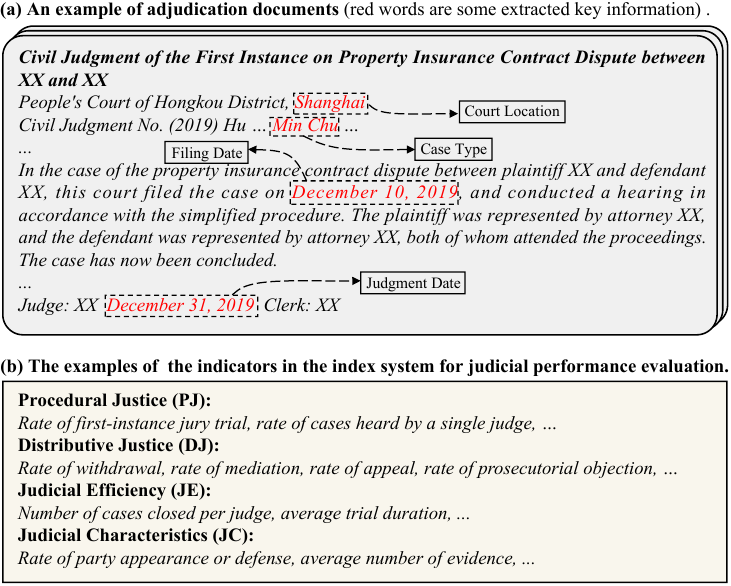}
    \caption{The examples of (a) adjudication documents and (b) judicial performance indicators.}
    \label{fig:dataprocessing}
\end{figure}
\section{Preliminaries} 


\subsection{Data Preparation}\label{sec:subsection3.1}  
This study utilizes China's judicial data and FDI data, covering the period from 2016 to 2019. The FDI data record annual foreign direct investments (FDIs) of all cities in China. Such data are extracted from China City Statistical Yearbooks, which are sourced from China Economic and Social Big Data Research Platform\footnote{\url{https://data.oversea.cnki.net/}}. The judicial data refer to over twelve millions of unstructured adjudication documents, sourced from China Judgments Online database\footnote{\url{https://wenshu.court.gov.cn/}}. 

In order to measure the judicial performance corresponding to each city, and further help predict city-level FDI, we design a comprehensive index system based on the given adjudication documents, guided by legal domain knowledge. This system comprises 380 indicators which are classified into four types: 1) procedural justice (PJ), reflecting the fairness of decision processes in adjudication~\cite{pj}; 2) distributive justice (DJ), indicating the fairness of decision outcome~\cite{pjdj}; 3) judicial efficiency (JE), measuring the efficiency of judicial decisions~\cite{je}; 4) judicial characteristics (JC), capturing other features of judicial performance. Figure~\ref{fig:dataprocessing}(b) exhibits several examples of these four types indicators.

To obtain the above designed judicial performance indicators, we extract and compute key information such as court location, case types, filing dates, and judgment dates from adjudication documents (illustrated in Figure~\ref{fig:dataprocessing}(a)). Then, according to the judgment dates and court locations, the case-level data are grouped into city-level data associated with each city in each year, so that a tabular dataset is built. In this dataset, each row contains two indexes which are a city and a specific year respectively, and the values of 380 judicial performance indicators. 

 \subsection{Problem Definition}
In this paper, city-level FDI prediction aims to predict the unknown FDI value of a given city in a specific year with its judicial performance indicators. We denote the judicial performance data as $\bm{X}=\{\bm{x}_1, \bm{x}_2, \dots, \bm{x}_{N}\}$, where $N$ means the number of all samples, $\bm{x}_i \in \mathbb{R}^{K}$ is the $i$-th sample denoting specific judicial performance indicators of a city in some year (e.g., $\langle$Shanghai, 2019$\rangle$), and ${K=380}$ denotes the number of indicators. The FDI data is denoted as $\bm{Y}=\{y_1, y_2, \dots, y_{N}\}$, where $y_i$ denotes the FDI of $\bm{x}_i$. $\bm{X}$ and $\bm{Y}$ compose the training data, which are utilized to learn a function $\mathcal{F}(\bm{X};\bm{Y};\bm{\theta})$ with all trainable parameters $\bm{\theta}$. When given a new sample with its judicial performance indicators, we use $\mathcal{F}$ to compute the corresponding FDI.
\section{Methodology}
\begin{figure*}[t]
    \centering
    \includegraphics[width=1\textwidth]{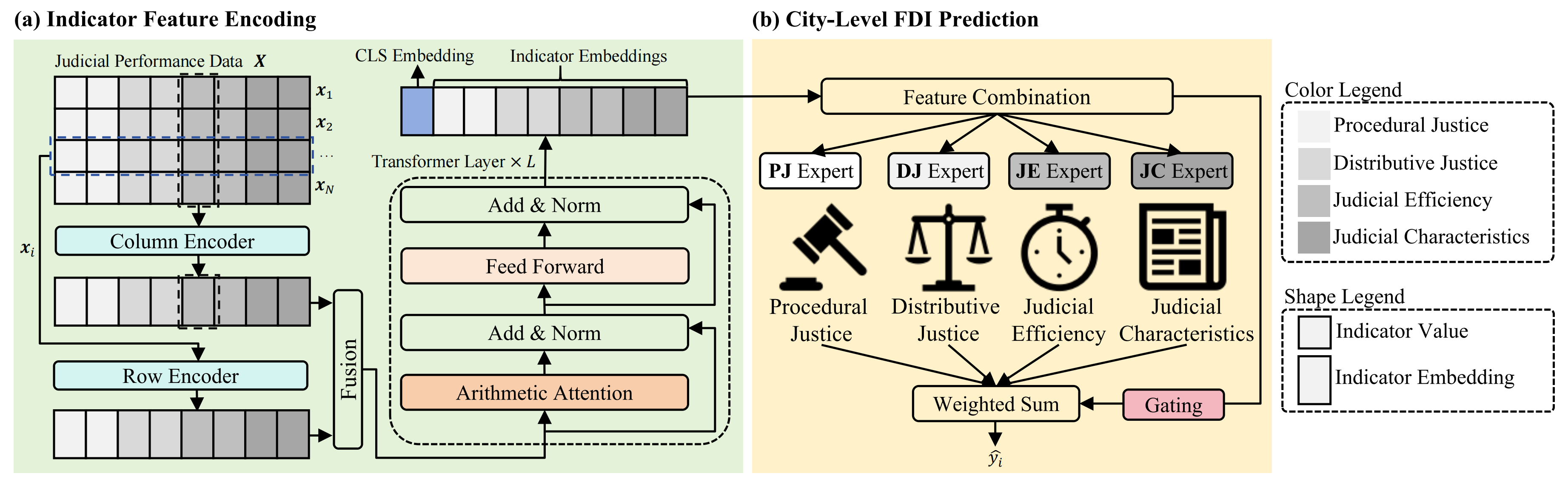}
    \caption{The framework of TLJD. (a) Indicator Feature Encoding maps each judicial performance indicator value to an embedding. (b) City-Level FDI Prediction leverages an MoE model consisting of four expert models to predict city-level FDI.}
    \label{fig:modelstructure}
\end{figure*}

In this section, we introduce our method TLJD for city-level FDI prediction in detail. As shown in Figure~\ref{fig:modelstructure}, TLJD consists of two main parts: 1) \textbf{Indicator Feature Encoding} which encodes values of judicial performance indicators as embeddings by transformer layers with arithmetic attention while simultaneously considering both row features and column features; 2) \textbf{City-Level FDI Prediction} which leverages an MoE model consisting of four expert models (i.e., PJ Expert, DJ Expert, JE Expert, and JC Expert) to predict city-level FDIs.

\subsection{Indicator Feature Encoding}
All judicial performance indicators in the built tabular dataset are numerical values, and TLJD transforms them into embeddings to capture complex relationships between such numerical features for downstream tasks, which is widely used in tabular learning~\cite{review2,2024review_tabular1}. More specifically, given the judicial performance data $\bm{X}$, we first design a row encoder to encode each row of indicators as a matrix describing each sample $\bm{x}_i$. We then propose a column encoder to map all columns of $\bm{X}$ to another vector space so that the global similarities between indicators can be computed. We finally fuse the output of these two encoders and apply transformer layers with arithmetic attention to incorporate the contextual information of relevant indicators for each indicator embedding. 

The row encoder uses $K$ different linear functions to encode $K$ indicators of each sample, respectively. For the $j$-th indicator $x_{i,j}$ of the sample $\bm{x}_i$, its encoded vector is denoted as $\phi^r_j(x_{i,j})$, where $\phi^r_j(x_{i,j}) = \bm{w}_jx_{i,j} + \bm{b}_j$ is the $j$-th linear function that maps the scaler value into a $d$-dimensional vector, $\bm{w}_j$ is a weight vector, and $\bm{b}_j$ is a bias vector.


For the indicators in the $j$-th column, we apply min-max scaling to normalize them to $[0,1]$, and denote the scaled indicators as $\bm{v}_j \in \mathbb{R}^N$. The column encoder applies two multi-layer perceptrons to encode $\bm{v}_j$ as follows:
\begin{equation}
\phi^c(\bm{v}_j) =
MLP_1
(\bm{v}_j)
\cdot 
MLP_2
(\bm{v}_j )
\end{equation}
where $MLP_1(\cdot)$ is a function that a two-layer fully-connected multi-layer perceptron (MLP) maps $\bm{v}_j$ to a $d$-dimensional vector, and $MLP_2(\cdot)$ is a function that the other two-layer fully-connected MLP transforms $\bm{v}_j$ into a scalar value to control the scale of the vector output by $MLP_1(\cdot)$.

By fusing the output of the row encoder and column encoder, the vector representation of the $j$-th indicator of $\bm{x}_i$ is computed as $\bm{h}_{i,j} =  \phi^r_j(x_{i,j})\odot \phi^c(\bm{v}_j) $, where $\odot$ denotes element-wise product. Then, a random initialized input CLS embedding $\bm{h}^\text{cls}$ and all $\bm{h}_{i,j}$ of $\bm{x}_i$ are concatenated horizontally in a matrix as
$\bm{H}_i= {[\bm{h}^{\text{cls}},\bm{h}_{i,1},\bm{h}_{i,2}, \dots, \bm{h}_{i,K}]}^\top $where $\top$ denotes matrix transpose. Here, $\bm{h}^\text{cls}$ is taken as trainable parameters and used for all samples. The setting of CLS embedding is to aggregate all indicator features of each sample.

With the output of the Fusion module $\bm{H}_i\in \mathbb{R}^{{(K+1)\times d}}$, we use transformer layers with arithmetic attention~\cite{2024amformer} to aggregate contextual information for each indicator embedding. In each transformer layer, the arithmetic attention contains additive and multiplicative attention operators, which are implemented based on the standard multi-head self-attention~\cite{vaswani2017attentionisallyouneed} as follows:
\begin{equation}
    MultiHead(\bm{H}_i)=[head_{1}, head_{2}, \ldots, head_{M}] \, \bm{W}^{O}
\end{equation}
\begin{equation}
    head_{m} = Attention(\bm{H}_i\bm{W}^Q_{m},\bm{H}_i\bm{W}^K_{m},\bm{H}_i\bm{W}^V_{m})
\end{equation}
\begin{equation}
    Attention(\bm{Q},\bm{ K}, \bm{V}) =Softmax(\frac{\bm{QK}^\top}{\sqrt{d^{'}}})\bm{V}
\end{equation}
where $\bm{W}^{O}\in \mathbb{R}^{d\times d}$ is a weight parameter, $M$ is the number of attention heads, $head_m\in \mathbb{R}^{d \times d^{\prime}}$ denotes the $m$-th attention head, ${d^{\prime}}=\frac{d}{M}$, $\bm{W}^Q_m$, $\bm{W}^K_m$, $\bm{W}^V_m \in \mathbb{R}^{d\times {d^{\prime}}}$ are trainable parameters for $head_m$, and $Softmax(\cdot)$ is an activation function converting a vector into a probability distribution with the same dimension.

Based on this, the additive attention operator is defined as:
\begin{equation}
    {Att}^{add}(\bm{H}_i)={MultiHead}(\bm{H}_i)
\end{equation}
while the multiplicative attention operator is defined as:
\begin{equation}
    {Att}^{mult}(\bm{H}_i)={exp}({MultiHead}({log}(Relu(\bm{H}_i)+\epsilon)))
\end{equation}
where ${exp}(\cdot)$ is the exponential function, ${log}(\cdot)$ is the logarithmic function, $Relu(\cdot)=max(0,\cdot)$ is an activation function, and $\epsilon$ is an all-one vector making each element of the input for ${log}(\cdot)$ greater than zero.
With both attention operators, the whole process of the arithmetic attention can be represented as:
\begin{equation} \label{att}
    Att(\bm{H}_i) =
    {FC}(
    [Att^{add}(\bm{H}_i)^\top,{Att}^{multi}(\bm{H}_i)^\top]
    )^\top 
\end{equation}
where $FC(\cdot)$ is a function that a fully connected layer 
transforms the $d\times2(K+1)$ matrix into a $d\times (K+1)$ matrix.




The transformer layer we used follows the structure in~\cite{vaswani2017attentionisallyouneed} with two sub-layers. The first one is the attention layer (we use the arithmetic attention in this paper) which outputs $Att(\bm{H}_i)$, and the second is a fully-connected feed-forward network. We employ residual connection for both sub-layers, followed by layer normalization. In this way, we define the output of the first transformer layer as $\bm{H}_i^{(1)}$, and the corresponding input is $\bm{H}_i$ denoted as $\bm{H}_i^{(0)}$, so the output $\bm{H}_i^{(l)}$ of the $l$-th transformer layer can be computed as follows:
\begin{equation}
    \bm{H}_i^{(l)} = TransLayer^{(l)}(\bm{H}_i^{(l-1)})
\end{equation}
where $TransLayer^{(l)}(\cdot)$ is a function representing all operations of the $l$-th transformer layer. Suppose we have $L$ transformer layers in total, the final encoded embeddings of $\bm{x}_i$ are denoted as $\bm{E}_i=\bm{H}_i^{(L)}$ composed of a CLS embedding and the embeddings of all indicators.

\subsection{City-Level FDI Prediction}
We apply an MoE model consisting of four expert models that focus on distinct aspects of judicial performance to predict city-level FDI. Given the output embeddings $\bm{E}_i$ of transformer layers, we first split them into four matrices, each of which is fed to the corresponding expert model to predict city-level FDI. We then use a gating network to dynamically compute the weights of expert models. Finally, the weighted sum of the output of all expert models is taken as the final result of city-level FDI prediction.

As mentioned in section \ref{sec:subsection3.1}, all judicial performance indicators are classified into four types, and the type set is represent as $\mathcal{T}=\{\text{PJ, DJ, JE, JC}\}$. In the module of Feature Combination, for a specific type $t\in \mathcal{T}$, we concatenate all indicator embeddings of $t$ and the CLS embedding from $\bm{E}_i$ to form a matrix $\bm{E}_{i}^t = {[\bm{e}_{i}^\text{cls}, \bm{e}_{i,1}^{t}, \bm{e}_{i,2}^{t}, \cdots, \bm{e}_{i,N^t}^{t}]}^\top
$, where $N^t$ denotes the indicator number of $t$,  $\bm{e}_{i,j}^{t}$ is the embedding of the $j$-th indicator of $t$, and $\bm{e}_{i}^\text{cls}$ is the CLS embedding of $\bm{E}_i$. As a result, the obtained matrices of four types are served as the input of our MoE model.



In the MoE model, four expert models are used to predict FDI, respectively. Each expert model corresponds to a matrix of specific indicator type. Given the matrix $\bm{E}_i^t$, the corresponding expert model applies a transformer layer following the same structure of the transformer layers used in Indicator Feature Encoding to aggregate contextual information for indicator embeddings and the CLS embedding within $\bm{E}_i^t$. The output matrix is denoted as ${\bm{E}_i^t}'=[{\bm{e}_{i}^{\text{cls}^t}}, {\bm{e}_{i,1}^{t'}}, {\bm{e}_{i,2}^{t'}}, \cdots, {\bm{e}_{i,N_t}^{t'}}]$, where ${\bm{e}_{i}^{\text{cls}^t}}$ is the output CLS embedding, ${\bm{e}_{i,j}^{t'}}$ is the output embedding of the $j$-th indicator of $t$. ${\bm{e}_{i}^{\text{cls}^t}}$ not only remains the general information of the sample $\bm{x}_i$, but also emphasizes the specific judicial performance information corresponding to the type $t$, which helps the expert model be the ``expert'' with indicators of $t$. Each expert model computes the city-level FDI $\hat{y}_{i}^{t}$ as follows:
\begin{equation}
     {\hat{y}_{i}^{t}} = {Linear}(
     {Relu}(
     {LayerNorm}(
     {\bm{e}_{i}^{{cls}^t}})))
\end{equation}
where $Linear(\cdot)=\bm{w}(\cdot) + \bm{b}$ is a linear function, $\bm{w}$ is the weight vector, $\bm{b}$ is the bias vector, and $LayerNorm(\cdot)$ is a function~\cite{lei2016layer} to normalize the input vector into a Gaussian distribution.



In the Gating module, we propose a gating network to generate the weight of each expert model for the final city-level FDI prediction as follows:

\begin{equation}
    a_i^t = \frac{exp(Gate(\bm{e}_{i}^{\text{cls}})^t)}{\sum_{t \in T} exp (Gate(\bm{e}_{i}^{\text{cls}})^t)}
\end{equation}
where $Gate (\cdot)$ is the gating network function and it is a fully connected layer transforming the $d$-dimensional vector into a four-dimensional vector, which corresponds to the four types in $\mathcal{T}$. $Gate (\cdot)^t$ represents the unnormalized expert model weight of the type $t$, and $a_i^t$ is the normalized weight. The final city-level FDI prediction result of $\bm{x}_i$ is computed as:
\begin{equation}
    \hat{y_i}=\sum_{t\in \mathcal{T}}a_i^t\hat{y}_i^t
\end{equation}
By dynamically adjusting the weights of expert models for different samples during prediction, the MoE model actually also adjust the weights of indicators which reveals their importance. Thus, we can utilize the weights of expert models for different cities to indicate the regional variations on how judicial performance influences city-level FDI.

\subsection{Training }
To minimize the errors between results of city-level FDI prediction and the ground truth, we use mean square error to compute the regression loss as follows: 
\begin{equation}
    \mathcal{L}_\textit{reg} = \frac{1}{N} \sum_{i=1}^{N} (\hat{y}_i - y_i)^2
\end{equation}

To encourage TLJD to model the judicial performance from different perspectives, we also use the expert responsibility loss~\cite{2022expertnet}~to optimize the prediction results of each expert model as follows:  

 
\begin{equation}
\mathcal{L}_{\text{er}} = -\frac{1}{N} \sum_{i=1}^N \log \sum_{t \in \mathcal{T}} a_i^t \exp \left(-\frac{(y_i - \hat{y}_i^t)^2}{2}\right)\label{equ_12}
\end{equation}
We train TLJD by minimizing $\mathcal{L}_\textit{reg}$ and $\mathcal{L}_\textit{er}$ jointly by stochastic gradient descent (SGD). The joint loss is defined as: 
\begin{equation}
    \mathcal{L}=
     (1-\lambda)\cdot
  \mathcal{L}_\textit{reg}+ \lambda\cdot\mathcal{L}_\textit{er}\label{equ_13}
\end{equation}
where $\lambda > 0$  is a hyper-parameter to balance $\mathcal{L}_\textit{reg}$ and $\mathcal{L}_\textit{er}$.
\begin{table}[t]
\centering
\captionsetup{justification=centering, singlelinecheck=false} 
\fontsize{8.5}{10}\selectfont
\begin{tabular}{lccccc} 
\toprule
Year        & 2016 & 2017 & 2018 & 2019 & ALL \\
\midrule
\# Document & 2.2M & 2.9M & 3.3M & 3.8M & 12.2M \\
\# Samples  & 265  & 265  & 267  & 263  & 1060  \\
\bottomrule
\end{tabular}
\caption{The statistics of collected adjudication documents and extracted samples for each year (M is short for million).}
\label{label:table2}
\end{table}

\begin{table*}[t]
\centering
\normalsize
\captionsetup{justification=centering,singlelinecheck=false}
\setlength{\tabcolsep}{4pt} 
\fontsize{8.5}{11}\selectfont
\begin{tabular}{c|ccc|ccc|ccc|ccc}
\toprule
                        & \multicolumn{3}{c|}{CCP (2016)} & \multicolumn{3}{c|}{CCP (2017)} & \multicolumn{3}{c|}{CCP (2018)} & \multicolumn{3}{c}{CCP (2019)} \\ 
\cmidrule(lr){2-4} \cmidrule(lr){5-7} \cmidrule(lr){8-10} \cmidrule(lr){11-13}
\multirow{-2}{*}{\cellcolor[HTML]{FFFFFF}Model}        & $R^2$ & RMSE  & MAE   & $R^2$ & RMSE  & MAE   & $R^2$ & RMSE  & MAE   & $R^2$ & RMSE  & MAE   \\
\midrule
\textbf{XGBoost}       & 0.4615      & 1.8186      & 0.7932      & \ul{0.5261}      & \textbf{1.4908} & 0.7649      & 0.5305      & 1.4877      & 0.7637      & 0.5129      & 1.4630      & 0.7101      \\
\textbf{LightGBM}      & 0.4037      & 1.8430      & 0.8515      & 0.4316           & 1.6858        & 0.8793      & 0.4152      & 1.6421      & 0.8756      & 0.5418      & 1.3946      & 0.7857      \\
\textbf{CatBoost}      & 0.5359 & 1.7237      & 0.6942      & 0.5203           & 1.6067        & \ul{0.6815 }     & \ul{0.5508} & \ul{1.4852} & 0.7952      & 0.5490      & 1.4613      & 0.6444      \\
\textbf{RandomForest}  & 0.3811      & 1.8946      & 0.7652      & 0.5129           & 1.6023        & 0.6881      & 0.4943      & 1.5646      & 0.7393      & 0.5621      & 1.3608      & 0.7203      \\
\midrule
\textbf{MLP}           & 0.3821      & 1.8564      & 0.8733      & 0.3963           & 1.7446        & 0.8688      & 0.3775      & 1.7414      & 0.9397      & 0.4069      & 1.4151      & 0.9008      \\
\textbf{ResNet}        & 0.3365      & 1.9379      & 0.8071      & 0.4322           & 1.6772        & 0.7404      & 0.4577      & 1.5858      & 0.8257      & 0.4754      & 1.5801      & 0.7744      \\
\midrule
\textbf{AutoInt}       & 0.4521      & 1.8364      & 0.7859      & 0.5136           & 1.5167        & 0.7908      & 0.4988      & 1.5582      & 0.8255      & 0.4413      & 1.5353      & 0.7349      \\
\textbf{SAINT}         & \ul{0.5432}      & 1.6903      & \ul{0.6834}      & 0.4832           & 1.6592        & 0.7235      & 0.4944      & 1.5402      & 0.7678      & \ul{0.6197} & \ul{1.3143} & 0.6578      \\
\textbf{FT-Transformer} & 0.4981     & 1.7315      & 0.7098      & 0.5103           & 1.5580        & 0.7064      & 0.4885      & 1.5562      & \ul{0.7124} & 0.5272      & 1.4318      & 0.6923      \\
\textbf{AMFormer}      & 0.5354      & \ul{1.6808} & 0.6864 & 0.4714           & 1.6233        & 0.7037 & 0.4755      & 1.5902      & 0.7257      & 0.6004      & 1.3153      & \textbf{0.5934} \\
\textbf{TLJD} (Ours)   & \textbf{0.5471} & \textbf{1.6768} & \textbf{0.6681} & \textbf{0.5324} & \ul{1.5077} & \textbf{0.6740} & \textbf{0.5565} & \textbf{1.4836} & \textbf{0.7083} & \textbf{0.6220} & \textbf{1.2868} & \ul{0.6054} \\
\bottomrule
\end{tabular}
\caption{The comparison results of CCP on four single-year datasets.}
\label{table:cc}
\end{table*}

\begin{table}[t]
\centering
\normalsize
\captionsetup{justification=centering,singlelinecheck=false} 
\setlength{\tabcolsep}{2.7pt} 
\fontsize{8.5}{11}\selectfont
\begin{tabular}{>{\columncolor[HTML]{FFFFFF}}c|ccc|ccc}
\toprule
\multicolumn{1}{c|}{} & \multicolumn{3}{c|}{CCP (mixed-year)} & \multicolumn{3}{c}{CTP} \\ 
\cmidrule(lr){2-4} \cmidrule(lr){5-7}
\rowcolor[HTML]{FFFFFF} 
\multirow{-2}{*}{\cellcolor[HTML]{FFFFFF}Model} & \multicolumn{1}{c}{$R^2$} & \multicolumn{1}{c}{RMSE} & \multicolumn{1}{c|}{MAE} & \multicolumn{1}{c}{$R^2$} & \multicolumn{1}{c}{RMSE} & \multicolumn{1}{c}{MAE} \\
\midrule
\textbf{XGBoost}      & 0.6844 & 1.4551 & 0.6119 & 0.6259 & 1.3621 & 0.8291 \\
\textbf{LightGBM}     & 0.7077 & 1.4003 & 0.6898 & 0.5431 & 1.5053 & 0.9163 \\
\textbf{CatBoost}     & 0.7158 & 1.3810 & 0.5827 & 0.6466 & 1.3240 & 0.8406 \\
\textbf{RandomForest} & 0.6768 & 1.5304 & 0.6768 & 0.6235 & 1.3675 & 0.8906 \\
\midrule
\textbf{MLP}          & 0.6507 & 1.5309 & 0.6370 & 0.4666 & 1.6264 & 0.6307 \\
\textbf{ResNet}       & 0.6664 & 1.4961 & 0.5634 & 0.4358 & 1.6273 & 0.5041 \\
\midrule
\textbf{AutoInt}      & 0.8655 & 0.9362 & 0.4453 & 0.9093 & 0.6704 & 0.3426 \\
\textbf{SAINT}& 0.7238 & 1.3612 & 0.5872 & 0.8667 & 0.8129 & \ul{0.3346} \\
\textbf{FT-Transformer} & \ul{0.8754} & \ul{0.9011} & \ul{0.4155} & 0.8855 & 0.7532 & 0.3574 \\
\textbf{AMFormer}& 0.8404 & 1.0201 & 0.4362 & \ul{0.9177} & \ul{0.6386} & 0.3603 \\
\textbf{TLJD} (Ours)  & \textbf{0.9217} & \textbf{0.7872} & \textbf{0.4137} & \textbf{0.9242} & \textbf{0.5626} & \textbf{0.3032} \\
\bottomrule
\end{tabular}
\caption{The comparison results of CCP on the mixed-year dataset and CTP.}
\label{table:cc_ct}
\end{table}


\section{Experiment}
\subsection{Experiment Settings}
\spara{Tasks and Datasets.}
City-level FDI prediction in real-world applications typically involves two application scenarios, i.e., cross-city prediction (CCP) and cross-time prediction (CTP), which are two evaluation tasks used in our experiments. CCP is to estimate missing historical FDIs for specific cities, and CTP is to predict future FDIs for the given cities. 

Based on our built tabular dataset and collected FDI data, we constructed different datasets for CCP and CTP, respectively. For CCP, we built four single-year datasets which are composed of the data in each year respectively, and a mixed-year dataset which contains all data in four years. These five datasets were split into training, validation, and test sets (the division ration is $3:1:1$), respectively. For CTP, we used all data to build a dataset where the data in the first three years are split into a training set and a validation set with a ratio of $3:1$, while the data in the last year were used as the test set. Table~\ref{label:table2} exhibits the numbers of collected adjudication documents and extracted samples for each year. 



\spara{Baselines.} We compared TLJD with the following advanced methods on tabular learning for city-level FDI prediction:
\begin{itemize}[topsep=5pt,leftmargin=1em]
\item \textbf{Tree-based models:} \textbf{Random Forest}~\cite{2001randomforest}~is an ensemble model using multiple decision trees to reduce overfitting, which is widely used in FDI prediction. \textbf{XGBoost}~\cite{chen2016xgboost} is a gradient boosting model which performs better than neural networks in many tabular learning tasks. \textbf{LightGBM}~\cite{ke2017lightgbm}~is a gradient boosting model with great efficiency. \textbf{CatBoost}~\cite{prokhorenkova2018catboost}~is an advanced gradient boosting model that uses oblivious decision trees and the ordered boosting algorithm.
\item \textbf{Classic neural network models:} \textbf{MLP} is a traditional neural network model always used for capturing nonlinear relationships of features in FDI prediction. \textbf{ResNet}~\cite{2016resnet}~enables training the deeper networks by skip connections, mitigating the vanishing gradient problem.
\item \textbf{Attention-based models:} \textbf{AutoInt}~\cite{song2019autoint}~is a representative attention-based model that encodes features into embeddings and utilizes the self-attention mechanism to learn high-order feature interactions. \textbf{SAINT}~\cite{2021saint} applies a hybrid attention mechanisms to boost tabular learning performance. \textbf{FT-Transformer}~\cite{2021fttransformer}~is a typical transformer-based tabular learning model which incorporates feature embeddings with contextual information. \textbf{AMFormer}~\cite{2024amformer}~designs a modified transformer architecture for more accurate sample separation of tabular data.  
\end{itemize}
\spara{Evaluation Metrics.}
We evaluated TLJD and baselines with the following metrics: 1) \textbf{Coefficient of Determination ($R^2$)} is the proportion of the variation in the dependent variable that is predictable from the independent variables~\cite{r2}. It measures how well prediction results approximate the ground truth FDIs, and an $R^2$ closer to one indicates more accurate prediction results; 2) \textbf{Root Mean Squared Error (RMSE)} is the quadratic mean of the differences between the predicted values and the ground truth. 3) \textbf{Mean Absolute Error (MAE)} is the mean of absolute differences between the predicted values and the ground truth.



\spara{Implementation.}
We adopted Adam optimizer~\cite{adam} for SGD. We selected the optimal hyper-parameters of TLJD on both tasks with different datasets via grid search on each validation set and chose the best TLJD based on MAE. For CCP on four single-year datasets, the optimal hyper-parameters of TLJD are as follows: the embedding size $d=96$, the number of transformer layers $L=2$, the loss weight $\lambda=0.4$, the learning rate: $0.0001$, the batch size: $32$, the number of epochs: $100$. For CCP on the mixed-year dataset and CTP, the optimal hyper-parameters of TLJD are the same as follows: the embedding size $d=96$, the number of transformer layers $L=3$, the loss weight $\lambda=0.6$, the learning rate: $0.001$, the batch size: $32$, the number of epochs: $50$. We implemented TLJD using PyTorch and all experiments were executed on an NVIDIA RTX 3090 GPU card (24 GB) of a 128 GB, 2.90 GHz Xeon server.

\subsection{Result Analysis}
\spara{Performance Comparison.}
We compared TLJD with ten baselines on all six datasets. Table~\ref{table:cc} shows the results of CCP on four single-year datasets. TLJD outperforms baselines in most situations, because TLJD uses two encoders to embed both row features and column features, and integrate them to our proposed MoE model focusing on different perspectives of specifically designed judicial performance indicators. We can also find that tree-based models demonstrate strong competitiveness because they have effective regularization techniques to avoid overfitting. Classic neural network models perform poorly because their structures can not effectively capture feature interactions in tabular learning. In contrast, attention-based models show good performance, as they can handle complex relationships between features and aggregate contextual information in embedding learning.

Table \ref{table:cc_ct} shows the comparison results of CCP on the mixed-year dataset and CTP. Compared with the ten baselines, TLJD provides the best results in all evaluation metrics on both tasks, further demonstrating the superiority of TLJD. The high accuracy achieved in CCP (0.9217 $R^2$) on the mixed-year dataset and CTP (0.9242 $R^2$) also highlights the effectiveness of judicial data for city-level FDI prediction. Meanwhile, attention-based models (including TLJD) significantly outperform other types of baselines, which reflects that leveraging attention mechanisms to aggregate contextual information for each indicator embedding when we have more training data is important to accurate city-level FDI prediction. 

\spara{Ablation Study.} To validate the contribution of key modules in TLJD, we constructed two variants of TLJD and conducted ablation experiments for CCP on the mixed year dataset and CTP, respectively. For the first variant, we replaced the MoE model with a single model utilizing a transformer layer for prediction and denoted it as w/o moe; for the other variant, we removed the column encoder and denoted this variant as w/o ce. Table~\ref{table:ablation} presents the ablation results. We found that our method outperforms both variants, which indicates that 1) the MoE model is important as it can dynamically adjust the weights of judicial performance indicators; 2) the column encoder is a key design in TLJD, as the global similarities between indicators can be captured in tabular learning.

\begin{table}[t]
\centering
\normalsize
\captionsetup{justification=centering,singlelinecheck=false} 
\setlength{\tabcolsep}{2.7pt} 
\fontsize{8.5}{11}\selectfont
\begin{tabular}{>{\centering\arraybackslash\columncolor[HTML]{FFFFFF}}c|ccc|ccc}
\toprule
\rowcolor[HTML]{FFFFFF} 
\multicolumn{1}{c|}{\cellcolor[HTML]{FFFFFF}} & \multicolumn{3}{c|}{\cellcolor[HTML]{FFFFFF}CCP (mixed-year)} & \multicolumn{3}{c}{\cellcolor[HTML]{FFFFFF}CTP} \\
\cmidrule(lr){2-4} \cmidrule(lr){5-7}
\rowcolor[HTML]{FFFFFF}
\multirow{-2}{*}{\cellcolor[HTML]{FFFFFF}Model} & \multicolumn{1}{c}{$R^2$} & \multicolumn{1}{c}{RMSE} & \multicolumn{1}{c|}{MAE} & \multicolumn{1}{c}{$R^2$} & \multicolumn{1}{c}{RMSE} & \multicolumn{1}{c}{MAE} \\ 
\midrule
\textbf{TLJD} (Ours) & \textbf{0.9217} & \textbf{0.7872} & \textbf{0.4137} & \textbf{0.9242} & \textbf{0.5626} & \textbf{0.3032} \\
\textbf{w/o moe}    & 0.8590       & 0.9726       & 0.5148       & 0.8251       & 0.8319       & 0.4644       \\

\textbf{w/o ce}     & 0.8821       & 0.8892       & 0.4698       & 0.8464       & 0.8723       & 0.4367       \\
\bottomrule
\end{tabular}
\caption{The results of ablation study for CCP on the mixed year dataset and CTP.}
\label{table:ablation}
\end{table}

\begin{figure}[t]
    \centering 
    \includegraphics[width=\linewidth]{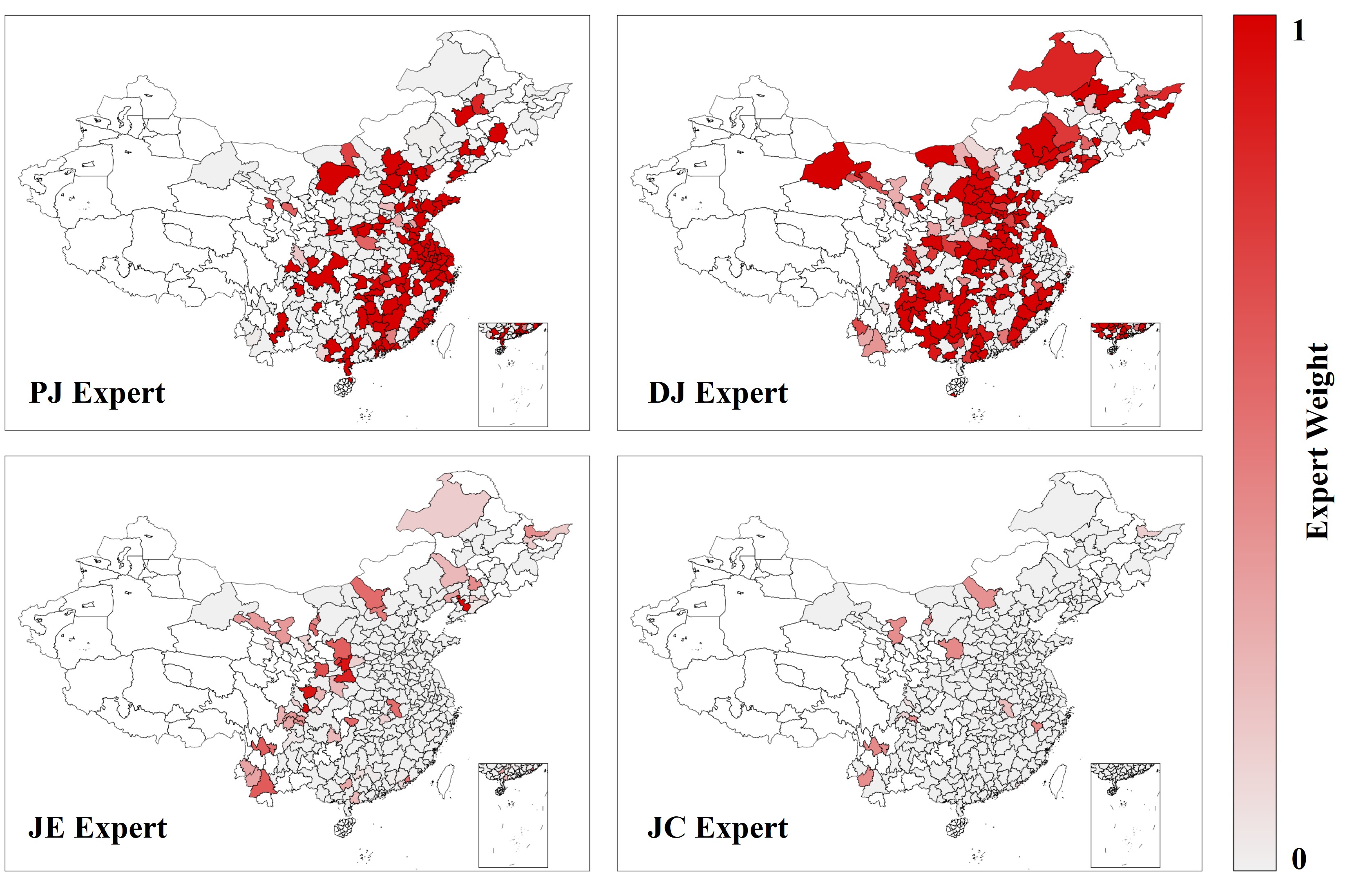}
    \caption{Regional variations of expert weights.}
    \label{fig:enter-label5} 
\end{figure}

\begin{figure}[t]
    \centering
    \begin{subfigure}[t]{0.45\linewidth}
        \centering
        \includegraphics[width=\linewidth]{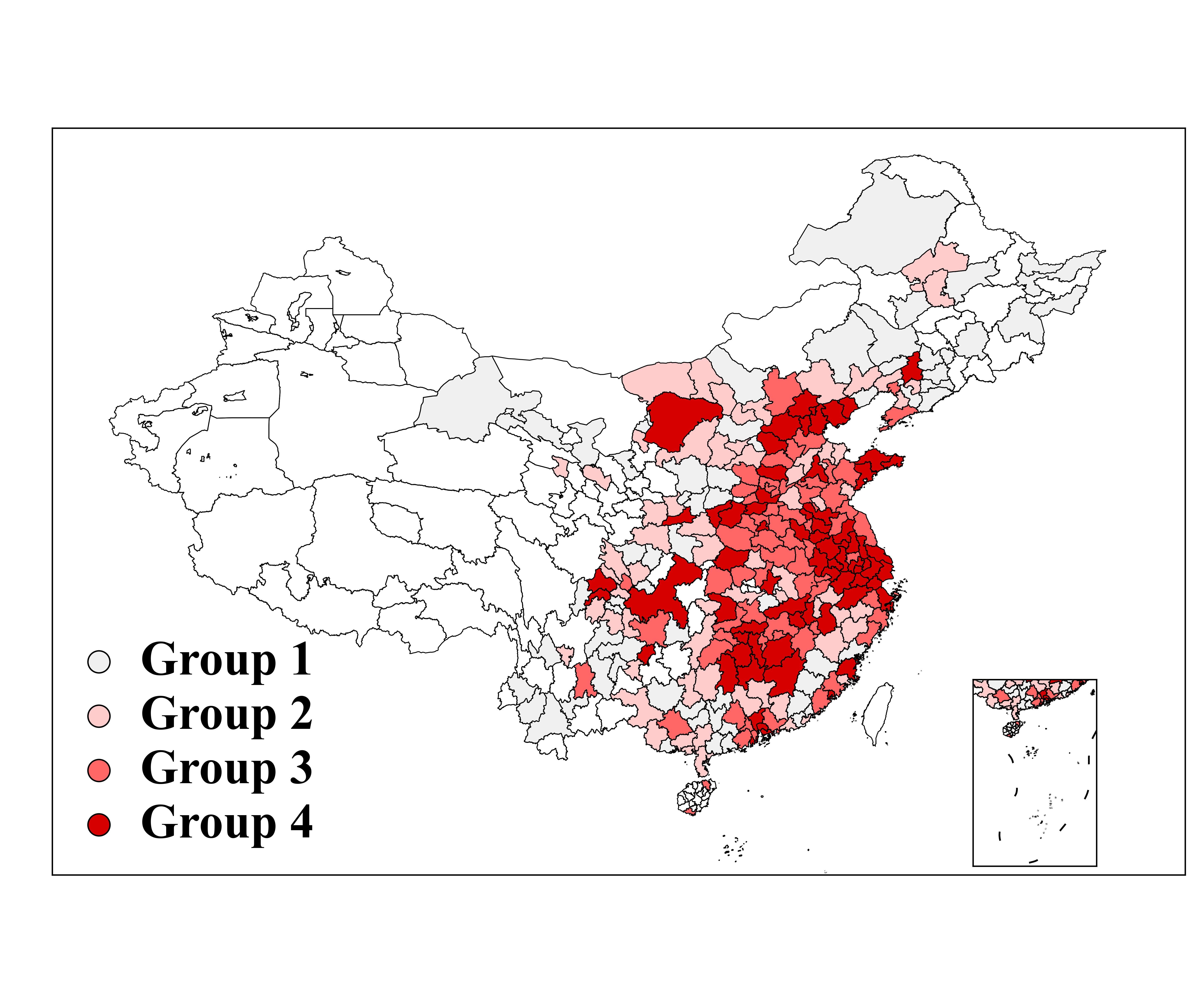}
        \caption{ }
        \label{fig:fdi_geo}
    \end{subfigure}
    \hfill
    \begin{subfigure}[t]{0.53\linewidth}
        \centering
        \includegraphics[width=\linewidth]{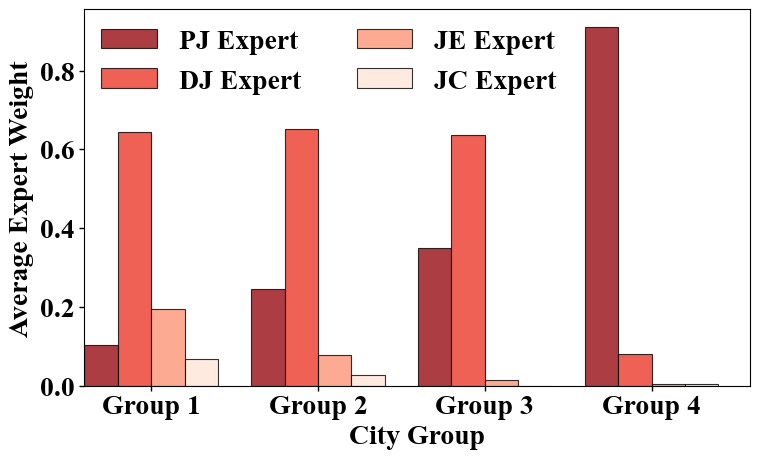}
        \caption{ }
        \label{fig:fdi_moe}
    \end{subfigure}
    \captionsetup{aboveskip=0pt, belowskip=0pt}
    \caption{(a) Regional variations of FDI; (b) Average expert weights of different groups of cities.}
    \label{fig:fdi-subplots}
\end{figure}

\spara{Expert Weight Analysis}.
To analyze the regional variations of expert (i.e., four expert models) weights and the association between the expert weights and city-level FDI, we conducted analysis on the test set (i.e., all data in 2019) of CTP. The results are shown in Figure~\ref{fig:enter-label5} and Figure~\ref{fig:fdi-subplots}, where the cities represented in white indicates that their data are unavailable. Figure~\ref{fig:enter-label5} visualizes the regional variations of expert weights, with the colors ranging from gray to red which indicate increasing expert weights. We can find that the PJ expert and DJ expert dominate the city-level FDI prediction in most cities, highlighting the significance of procedural justice indicators and distributive justice indicators which influence the decisions of foreign investors.
In particular, the PJ expert exhibits larger weights in China's eastern coastal regions, whereas the DJ expert has a stronger influence in the central, western, and northeastern regions. This may be driven by differences in regional socio-economic conditions, such as disparities in economic development and local policy priorities.

To further analyze the association between expert weights and city-level FDI, we divide all cities into four groups. We ranked all cities by FDI in descending order, then group 1 (the last 25\% cities), group 2 (the top 50\%$\sim$75\% cities), group 3 (the top 25\%$\sim$50\% cities), and group 4 (the top 25\% cities) were created. Figure~\ref{fig:fdi-subplots}(a) shows the distribution of cities within the four groups, and a darker color illustrates a larger FDI. The average expert weights for each group are visualized in Figure~\ref{fig:fdi-subplots}(b). We can see that the cities with lower FDIs (Group 1, 2, 3) have a larger average weights of the DJ expert, while the cities with higher FDIs (Group 4) have a much larger average weight of the PJ expert. These mean that in the regions with a well-established investment environment (i.e., the regions with high FDIs), procedural justice is more important and investors prioritize the fairness and transparency of judicial procedures. However, in the regions with a emerging or less-developed investment environment (i.e., the regions with low FDIs), distributive justice is more important and investors place greater emphasis on the practical outcomes of judicial decisions, which determines whether the judiciary can effectively solve the problems of investors.


\section{Conclusion}

In this paper, we propose to use judicial data to predict city-level FDI with a new tabular learning method TLJD, which addresses the issue of relying on the economic data prone to manipulation, and can support economic decision making of local governments. Experimental results show the superiority of TLJD compared with different baselines, and the effectiveness of key modules in the ablation study. By providing a new technical path of city-level FDI prediction, our study has potential applications not only in China but also in other major global economies, which contributes to economic growth and the achievement of sustainable development goals. 

\section*{Acknowledgements}
This work is supported by the National Natural Science Foundation of China (Grant No. 62376058, 62302095), the National Social Science Fund of China (Grant No. 21CFX025), the Southeast University Interdisciplinary Research Program for Young Scholars, and the Big Data Computing Center of Southeast University.

\section*{Contribution Statement}
This study was finished under the cross-disciplinary collaboration between the AI research team led by Prof. Tianxing Wu, and the law and economics research team led by Prof. Yuqing Feng. Both teams contributed equally.


\bibliographystyle{named}
\bibliography{ijcai25(3)}

\end{document}